\documentclass[manuscript]{acmart}
\AtBeginDocument{%
  \providecommand\BibTeX{{%
    \normalfont B\kern-0.5em{\scshape i\kern-0.25em b}\kern-0.8em\TeX}}}

\usepackage{lipsum}
\usepackage{xspace}
\usepackage{arydshln}

\newcommand{\bertmodel}{ViralBERT\xspace}
\DeclareCaptionLabelFormat{andtable}{#1~#2  \&  \tablename~\thetable}

\copyrightyear{2022} 
\acmYear{2022} 
\setcopyright{rightsretained} 
\acmConference[UMAP '22 Adjunct]{Adjunct Proceedings of the 30th ACM Conference on User Modeling, Adaptation and Personalization}{July 4--7, 2022}{Barcelona, Spain}
\acmBooktitle{Adjunct Proceedings of the 30th ACM Conference on User Modeling, Adaptation and Personalization (UMAP '22 Adjunct), July 4--7, 2022, Barcelona, Spain}
\acmDOI{10.1145/3511047.3536415}
\acmISBN{978-1-4503-9232-7/22/07}

\begin{document}

\title{\bertmodel: A User Focused BERT-Based Approach to Virality Prediction}


\author{Rikaz Rameez}
\affiliation{%
  \institution{University College London}
  \streetaddress{Gower Street}
  \city{London}
  \country{UK}
  \postcode{WC1E 6BT}}
\email{rikaz.rameez.18@ucl.ac.uk}

\author{Hossein A.~Rahmani}
\affiliation{%
  \institution{University College London}
  \streetaddress{Gower Street}
  \city{London}
  \country{UK}
  \postcode{WC1E 6BT}}
\email{h.rahmani@ucl.ac.uk}

\author{Emine Yilmaz}
\affiliation{%
  \institution{University College London}
  \streetaddress{Gower Street}
  \city{London}
  \country{UK}
  \postcode{WC1E 6BT}}
\email{emine.yilmaz@ucl.ac.uk}


\begin{abstract}
Recently, Twitter has become the social network of choice for sharing and spreading information to a multitude of users through posts called `tweets'. Users can easily re-share these posts to other users through `retweets', which allow information to cascade to many more users, increasing its outreach. Clearly, being able to know the extent to which a post can be retweeted has great value in advertising, influencing and other such campaigns. In this paper we propose \bertmodel, which can be used to predict the virality of tweets using content- and user-based features. We employ a method of concatenating numerical features such as hashtags and follower numbers to tweet text, and utilise two BERT modules: one for semantic representation of the combined text and numerical features, and another module purely for sentiment analysis of text, as both the information within text and it's ability to elicit an emotional response play a part in retweet proneness. We collect a dataset of 330k tweets to train \bertmodel and validate the efficacy of our model using baselines from current studies in this field. Our experiments show that our approach outperforms these baselines, with a $13\%$ increase in both F1 Score and Accuracy compared to the best performing baseline method. We then undergo an ablation study to investigate the importance of chosen features, finding that text sentiment and follower counts, and to a lesser extent mentions and following counts, are the strongest features for the model, and that hashtag counts are detrimental to the model.
\end{abstract}

\maketitle

\section{Introduction}
\label{sec:intro}
The zeitgeist of recent times has been characterised by the rise of the internet as a central component in our lives, and a huge factor in this is the communication between users on social media allowing borderless conversations between groups of people. Twitter is one of the biggest social media platforms in the world with over 300 million monthly active users. These users can share `tweets' - a piece of text up to 280 characters in length with the option to be accompanied with photos, GIFs or videos. Around 500 million tweets are posted every day on the platform, so there is a huge amount of content and interaction between users. Users can interact with tweets by sharing the post to their follower network through retweets, liking them, or replying to the post with their own tweet. These interactions show that a tweet contains information that users have interest in and by retweeting, this information gets spread to other users. A small number of tweets get more interactions than others, so gain attention from a relatively large number of users on Twitter - this indicates that the tweet has gone viral. Virality can be used to determine popularity and engagement of trends and topics not only on Twitter but throughout society as a whole. Indeed, there has been an increasing number of studies on the phenomenon of viral posts affecting social, economic and political outcomes so there is importance in having the capability to forecast the popularity of a tweet \cite{Borges2019}. \citet{Kouzy2020} found that false or unverifiable content about the COVID-19 pandemic could just as easily gain a high amount of user interactions as verified information, showing the extent to which misinformation can rampantly spread on social media (as it did in the pandemic \cite{Volkmer2021}). In addition, being able to model Twitter popularity would allow tweets to be engineered towards user engagement, which has uses in viral marketing or public information campaigns for example.
There are analogous popularity metrics on other social media (Likes/Shares on Facebook, Instagram, YouTube etc.), so this study could also ignite research into these platforms.

A large amount of current studies on tweet engagement focus on modelling the network structure of Twitter and how tweets propagate between users and demographics \cite{Weng2013, Lympero2021}. This allows some understanding of Twitter's userbase and post structure and why users share posts between each other. However, there is limited recent research on predicting virality or popularity, and these studies focus on a specific subset of tweets \cite{Zafra2021} or certain Twitter users \cite{Pancer2016} instead of tweets from the entire userbase. Focusing on a specific subset of tweets or users allows adequate performance on a minority of tweets, but no generalisation to Twitter users and posts as a whole. Predicting virality poses a challenge as it can be affected by a multitude of factors, many of which are not easily quantifiable such as relatability to users, creativity of content and relevancy to current social climates - modelling this widespread user appeal is a difficult problem. Furthermore, most tweets are never retweeted, so it is also difficult to build a large, balanced and representative dataset on all Twitter users for ML use due to zero inflation. In this paper, we build our own dataset of a sample of 330k tweets to train our model. We then explore whether we can predict the degree of virality of a tweet using novel deep leaning methods on this problem, namely using the BERT architecture \cite{bert}. We create a well performing virality prediction model using textual, content-based, and user-based features from a tweet, called \bertmodel. Our model incorporates a specific BERT module for sentiment of tweet text, as emotional arousal has been highly linked with virality in past works \cite{Berger2012}, and this correlation is further investigated in this paper. \bertmodel is compared with models in the most recent work on predicting Twitter virality, and the importance of features is investigated as well.
\section{Background}
\label{sec:background}
There are several studies on analyzing the trends of and modelling user interactions on Twitter. \citet{Lympero2021} investigates the temporal growth of retweet numbers and identifies a pattern of virality evolution for all tweets - they start with rapid growth and then asymptotically reach a virality limit after a certain amount of time. \citet{Kwak2010} explores several different ways of modelling Twitter usage. The Pagerank score of a user is compared to the amount of followers that a user has and the amount of retweets a user gets.
It is found that Pagerank is similar to the follower amounts but having a large amount of followers doesn't necessarily mean that a tweet from this user will have a high amount of retweets. Furthermore, the paper emphasises that retweets are important for user exposure - even if a user doesn't have a lot of followers, being retweeted will give them exposure to a large amount of users.

There have also been some studies into modelling tweets with the aim of predicting if a post will be popular. \citet{Zafra2021} focuses on virality depending on sentiments (and other features) during the Catalan referendum. 
They find that likes, hashtags, URLs, mentions, followers, and following numbers have significant effects on retweet numbers, and reply numbers and verification status have negative effects on retweet numbers. Notably, they also find that positive terms contribute negatively to retweets count and negative terms contribute positively - similar conclusions have been made by \citet{Pancer2016}. \citet{Nesi2018} assess the degree of retweeting, and treat this problem as a multi-class classification problem using retweet class boundaries (0, 1-10, 11-100, 101-1000, 1000+). The features used are the hashtags, mentions and URLs in a tweet, the number of favourites, and tweet author's followers and following numbers. The models used on these features are Random Forests and their own model, RPART, which is form of decision tree. They find that mentions count is the most important metric in predicting virality degree. An observation from this study is that favourites count is used, but as it is an indicator of user engagement it should not have been used.

These studies provide insight into methodologies that could be used in this study. The key observations for this study are the features that could be used in addition to tweet text, such as content and author based features. In addition, retweets are used as the principal indicator of virality as retweets increase user visibility of a tweet and are driven mainly by tweet content, so this should inform our labels. Virality prediction could also be turned into a classification problem, which can allow for multiple tasks to be used to inform a prediction, and allow for different classification metrics to be used. Although ML methods have been used in previous works, there is a clear lack of studies in using deep learning algorithms and NLP techniques to predict tweet virality - this will be the focus and novelty of this work.
\section{Proposed Methodology}
\label{sec:method}

\begin{figure}
    \includegraphics[width=0.47\textwidth]{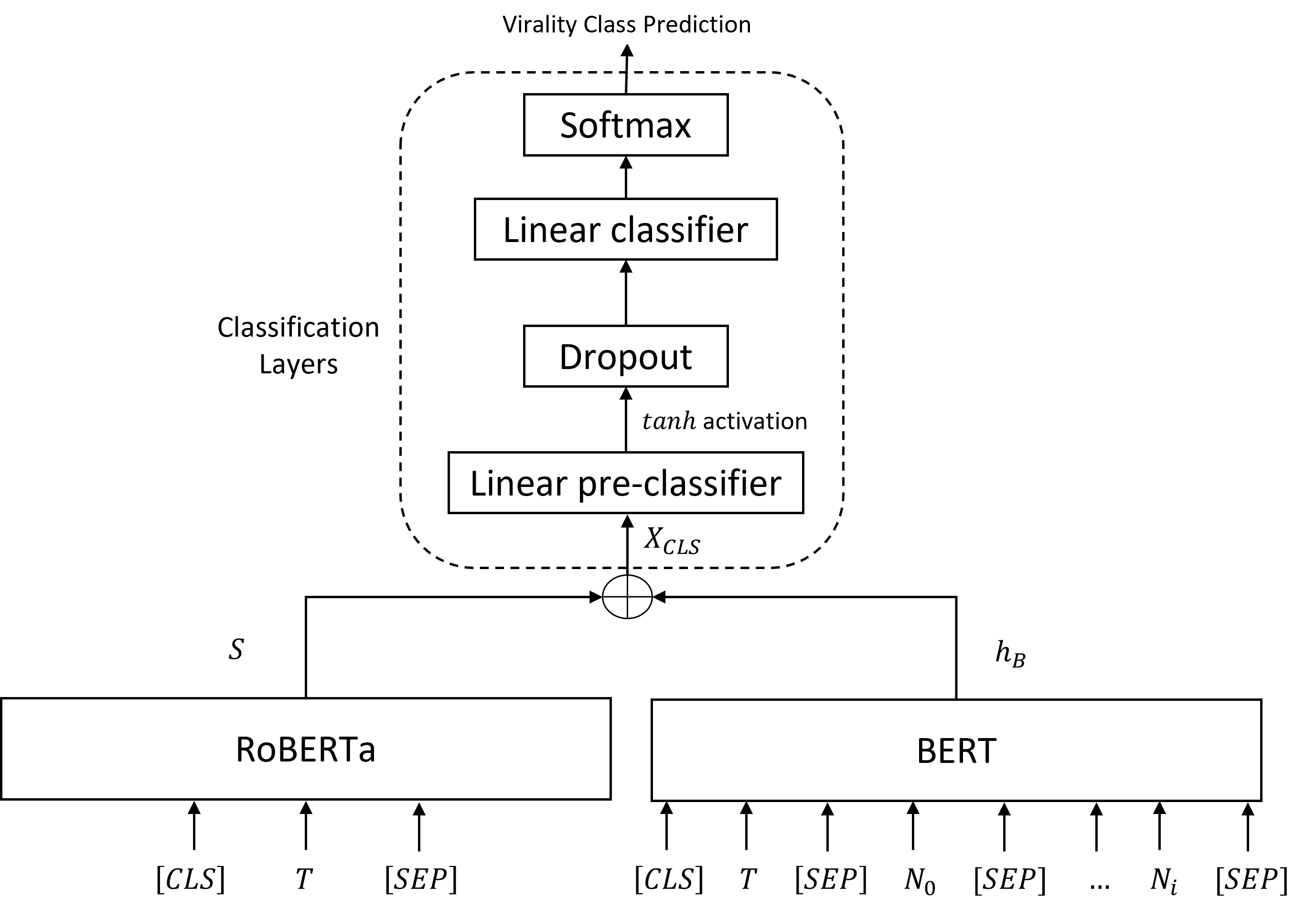}
    \Description{Architecture using tweet text and numerical features for BERTweet, then concatenating the output with RoBERTa sentiment analysis for classification layers}
    \caption{\bertmodel architecture 
    }
    \label{fig:arch}
\end{figure}

In this section, we describe our proposed approach for virality prediction, called \bertmodel, which captures both numerical and textual information from tweets.
The architecture is shown in Figure \ref{fig:arch}. Our model generates a high dimensional representation using a combination of tweet text and numerical features, as well as a probability distribution of the sentiment of tweet text, then feeds this into a classification model to produce a virality class prediction. Sentiment is used here as several studies \cite{Zafra2021,Pancer2016} find it directly affects the propagation of a tweet. Therefore, we use BERT \cite{bert} on the tweet text for both this sentiment task and tweet text semantics task - BERT is a bidirectional transformer based language model that is the current state-of-the-art for NLP models. The BERT modules we use are detailed below.

\subsection{BERT}
Tweets, due to their short length, tend to use informal grammar and shortened vocabulary (e.g hashtags, abbreviations, etc.) so there is a challenge in choosing a pre-trained model adapted to this domain as opposed to the general domain such as Wikipedia.
The pre-trained model we use is BERTweet \cite{bertweet} which is already fine-tuned on a corpus of 850M tweets, meaning that it has a deep understanding of language used in tweets from a wide range of topics and users.
Indeed, BERTweet's vocabulary contains many token mappings that wouldn't be in a conventional BERT vocabulary such as emojis and colloquialisms, so BERTweet's specific tokeniser was used.
BERTweet is based on BERT base, with 12 transformer blocks, 768 hidden units and 12 self-attention heads and produces a 768 dimension $[CLS]$ embedding.
The input into this part of the model is the tweet text, $T$, concatenated with the non-normalised numerical features ($N_{0}, N_{1} ... , N_{i}$), which are specified in Section \ref{m-trainingparams}.
This is separated using BERT's $[SEP]$ tokens to indicate multiple sentences for each feature, so during training it is being fine-tuned to learn virality from these features concatenated as sentences.
Formally, the input to BERTweet is as follows, and the output produced afterwards is a $[CLS]$ embedding, $h_B$.

\begin{equation}
    X_{BERT} = [CLS] \oplus T \oplus [SEP] \oplus N_{0} \oplus [SEP] \oplus N_{1} \oplus [SEP] \oplus ... \oplus N_{i} \oplus [SEP]
\end{equation}

\subsection{RoBERTa}
For sentiment analysis we use a RoBERTa \cite{roberta} based pre-trained model.
This is an optimised method for pre-training BERT where key hyperparameters are modified such as removing the Next Sentence Prediction objective and training with larger mini-batches. This model is from the TweetEval benchmark \cite{cardiffnlp} so is also specific to tweet texts. The output of this model is a likelihood for a tweet having each of negative, neutral or positive sentiment so is a 3 dimensional softmaxed probability distribution, which we call $S$. Sentiment analysis is used as a task alongside BERTweet so it is an additional numerical input for the final virality classification layers and it is fine-tuned for this task during training.

\subsection{Classification Layers}
The output of BERTweet, $h_B$, and sentiment probabilities, $S$, is concatenated as input into the classification layer:
\begin{equation} \label{eq:txt-cls}
    X_{CLS} =  BERTweet(X_{BERT}) \oplus S = h_B \oplus S
\end{equation}
BERT is fine-tuned for virality prediction using this layer - using fine-tuning allows ease of implementation, faster training and performance gains as we build on pre-trained models \cite{Huang2019}. The classification layers we use has a fully connected flat MLP structure with tanh activation - the dimensions of this layer are based on $X_{CLS}$, so there are 771 hidden units. Next, a dropout of 0.1 is applied before a final linear prediction layer, outputting probabilities of belonging to a virality class after being fed into a softmax function. Experimentation was done on increasing the amount of layers for the classifier and using CNNs instead of ANNs, but performance was not improved by either.
\section{Experimental Setup}
\label{sec:experiments}

\subsection{Data Collection}
The datasets for this study are collected through the Twitter API v2\footnote{\url{https://developer.twitter.com/en/docs/twitter-api}} using Python. The Twitter API allows context annotations\footnote{\url{https://developer.twitter.com/en/docs/twitter-api/annotations/overview}} to be used to extract tweets that fall under a certain context, categorised by domains and entities. For this study, we use 8 domain entity pairings - Cryptocurrencies, TV \& Movies, Pets, Video Games, Cell Phones, COVID-19, Football and K-pop. These topics are chosen as they are generally popular over time so would have consistent virality criteria during data collection, and are a good representation of a large amount of Twitter posts. Using these topics allow us to filter a large amount of the spam posts that are present on Twitter, and any that remain would be in the lower virality classes so do not affect the model as much. For this study, only original (not retweeted) English tweets are collected. These tweet fields we collect are text, creation time, number of hashtags, number of mentions, and tweet source client. 
Fields collected from the user are followers and following count, and verified status. Retweet, like, reply and quote counts are retrieved after 24 hours of creation as by this time the virality limit of a tweet would have been reached \cite{Chen2018,Lympero2021}, and these metrics are collected to indicate virality. This information is all connected using the tweet's ID and duplicate IDs are removed. The original dataset consists of 2.1 million tweets collected over 26 days, although this was reduced as detailed in Section \ref{m-exploration}. 

\subsection{Data Exploration and Labelling}
\label{m-exploration}
Data analysis is conducted on the data to ensure data was appropriate for modelling and to explore trends within the data. Within the original dataset over ninety percent of the tweets are tweets that have not been retweeted, which is indicative of how actual users interact with tweets. However, this skewed class distribution will lead to poor performance for an ML model. Therefore, the dataset is made more balanced (although not completely balanced) by reducing the number of tweets with zero retweets to be equal the amount with at least one retweet. The total amount of tweets remaining after this reduction is 330k with half of the tweets having zero retweets.

We decide that retweets are best used as the label as it is more indicative of tweet propagation to other users through social cascades \cite{Cheung2017,Zafra2021}, so provide more exposure and engagement than likes \cite{Pancer2016}. We use retweet class bands to turn the problem into a multi-class classification problem similar to \citet{Nesi2018}. 
Using classes instead of raw retweet numbers allows the problem to be translated to other platforms and allows the exploration of additional virality metrics for labelling in the future.
We select the following four class boundaries for virality degree: 0 retweets (164k tweets); 1 retweet (98k tweets); 2-20 retweets (60k tweets); 21+ retweets (5k tweets). From the tweet numbers in each class it is evident that there is still significant class imbalance present.

\subsection{Training Parameters}
\label{m-trainingparams}
We use Class-Balanced Focal Loss \cite{Cui2019} as the loss function to aid in handling the class imbalance (as half of the dataset has zero retweets).
This calculates loss with an additional class balancing factor, which is especially useful for long-tailed datasets such as ours \cite{Huang2021}. We apply AdamW \cite{AdamW} as the optimization algorithm to update model parameters and to allow for better generalisation. The numerical features chosen for all models are hashtags, mentions, followers, following, verified status, and text length. `Hashtags' and `mentions' are the amount of these elements in the tweet text, so is a feature derived from tweet text. In addition, the sentiment of tweet text is added as a feature for baselines. Tweet text embeddings are not used as none of the literature we review utilises them. Note that all of these features are collected when the tweet is collected and would stay consistent. With the exception of \bertmodel, a min-max scalar is used for the numerical features as there is a large variation in the values (e.g. hashtags being much lower than follower counts). The ratio for training, validation and test sets is 80:10:10, with data being split randomly into these subsets. A data loader class with a batch size of 32 is used, and models are trained until convergence on the validation set.

\subsection{Baselines}
\label{baseline}
We compare the performance of \bertmodel with the following baseline methods that were developed for similar tasks in the literature. Their inputs are numerical features as detailed in Section \ref{m-trainingparams} and are built using \citet{scikit-learn}:

\begin{itemize}
    \item Logistic Regression \cite{Hong2011}: Utilises Newton's method for gradient optimisation. This method has been used for predicting popular messages in Twitter \cite{Hong2011}.
    \item Support Vector Machine (SVM) \cite{Ma2013,Nesi2018}: Uses hinge loss and SGD optimization. This method used for predicting the popularity of newly emerging hashtags in Twitter \cite{Ma2013}, as well as for assessing the retweet proneness \cite{Nesi2018}.
    \item Decision Tree \cite{Nesi2018}: Uses Gini impurity score to measure quality of a split with no max depth. This method has also been used for assessing retweet proneness \cite{Nesi2018}.
    \item Random Forest \cite{Palovics2013, Nesi2018}: Uses 100 trees with no max depths. This baseline is based on previous work, which focused on the problem of temporal prediction of retweet count \cite{Palovics2013} and likelihood of a retweet \cite{Nesi2018}.
\end{itemize}

Two more baselines are added using either numerical or text features only to evaluate how these features perform.

\begin{itemize}
    \item MLP$_{Num}$: Fully connected Multi Layer Perceptron for numerical features. Has one hidden layer with 32 units.
    \item \bertmodel$_{Text}$: Using BERTweet on tweet text only (no numerical features)
\end{itemize}
To foster the reproducibility of our study, we have made the code for all models including \bertmodel open source. \footnote{\url{https://github.com/RikazR/ViralBERT}}

\subsection{Evaluation Metrics}
To evaluate the effectiveness of our classifiers we employ the following metrics on a holdout test set after training: accuracy, which is the ratio of correct predictions over total number of instances evaluated; recall, which is the fraction of actual classifications for a class that were correctly identified; precision, which is the proportion of model predictions for a class that were actually correct; F1 Score, which is the harmonic mean of precision and recall. As accuracy and F1 scores consider both the positive and negative classifications for evaluation, these are ideal for discriminating our optimal solution. These metrics are balanced using the macro weighting, so there is an equal weighting and importance for each class allowing metrics to be more indicative of predictions of all classes \cite{Grandini2020}.
\section{Results} 
\label{sec:results}

\begin{table}
\centering
\caption{Evaluation metrics on models. The best performing results are shown in \textbf{bold}.}
  \begin{tabular}{lcccc}
    \toprule
    Method & F1 Score & Precision & Recall & Accuracy \\
    \midrule
    Logistic Regression & 0.235 & 0.503 & 0.277 & 0.277  \\
    SVM & 0.221 & 0.320 & 0.271 & 0.271 \\
    Decision Tree Classifier & 0.405 & 0.402 & 0.408 & 0.408 \\
    Random Forest Classifier & 0.458 & 0.562 & 0.435 & 0.435 \\ \hdashline
    MLP$_{Num}$ & 0.213 & 0.235 & 0.268 & 0.268 \\
    \bertmodel$_{Text}$ & 0.410 & 0.415 & 0.409 & 0.409 \\
    \midrule
    \bertmodel & \textbf{0.523} & \textbf{0.609} & \textbf{0.494} & \textbf{0.494} \\
  \bottomrule
\end{tabular}
\label{tab:results}
\end{table}

Table \ref{tab:results} shows the results for our different model architectures. 
Solely utilizing \bertmodel with text does not give optimal performance compared to the baseline models.
There is a significant performance increase when adding numerical features to \bertmodel, in comparison to only using text so numerical features are also important for virality prediction.
By fine-tuning \bertmodel to the concatenated input of numerical features, along with training classification layers, we are able to achieve evaluation metrics that are higher than the baselines.
This could mean that adding numerical features to text allows \bertmodel to gain context into features that could help with prediction in addition to using the semantics of the text input.

\begin{table}
\caption{Comparison of removing features on performance of \bertmodel. The most significant differences are shown in \textbf{bold}.}
\centering
  \begin{tabular}{lcccc}
    \toprule
    Feature removed & F1 Score & Accuracy \\
    \midrule
    \bertmodel & 0.523 & 0.494 \\
    \midrule
    Sentiment & \textbf{0.438} & \textbf{0.432} \\
    Hashtags & \textbf{0.531} & \textbf{0.502} \\
    Mentions & 0.490 & 0.466 \\
    Followers & \textbf{0.429} & \textbf{0.435} \\
    Following & 0.502 & 0.474 \\
    Verified & 0.518 & 0.491 \\
    Text Length & 0.523 & 0.493 \\
  \bottomrule
\end{tabular}
\label{tab:ablation}
\end{table}

We conduct a study to measure the importance of each feature to \bertmodel by removing a feature from the input and measuring the performance of the model. For sentiment, the RoBERTa sentiment module is removed.
Table \ref{tab:ablation} shows the results of removing features from the input in comparison to \bertmodel.
Firstly, removing either sentiment or follower counts completely from the network reduces the performance of the model significantly compared to other features.
This is intuitive, as a user with a greater amount of followers is likely to gain more traction on their tweets \cite{Hong2011}, and tweets that induce a greater emotional response in the users reading them may get retweeted more \cite{Hemsley2018}. Tweet sentiments allow the model to detect viral tweets from less popular users, and using both of these user-focused features will give the best model performance.
Furthermore, removing mentions count or following counts have an effect on performance, albeit lesser than the former two features. This could be due to more popular users tending to follow others less (to have a high follower-following ratio). Also, tweets having lots of mentions lack readability and utilise valuable space that could be used for information that is useful to users, meaning it is less likely to be retweeted.
The most surprising result is that removing hashtag counts slightly improves the performance of the model. This could mean that this count is unrepresentative of virality, so adding this feature makes the BERT learn worse representations from the inputs.
These results should be explored further in future work with a larger dataset and a more comprehensive ablation study, including testing the interaction between each feature, to allow for a deeper understanding of the reasons these have an adverse effect on performance.

\section{Conclusion and Future Work}
\label{sec:conclusion}

In this paper, we propose a BERT-based approach for predicting tweet virality using both tweet text and numerical features, called \bertmodel. These numerical features contain both content based features such as hashtag counts and user based features such as author follower count. We collect a dataset of tweets from the general userbase of Twitter, and train our proposed models on this dataset, showing that our BERT-based approach outperforms current approaches to virality prediction with an F1 score of 0.523 and Accuracy of $0.494$. We find that text sentiment and follower counts, and to a lesser extent mentions and following counts, are the most important features in determining tweet virality, and that hashtag counts negatively affect the model performance.

Future work can be done on increasing the complexity of the proposed approaches and collecting a more balanced dataset, which was one of the challenges of this study. Methods of balancing the class imbalance present in the dataset could be used, for example using a hybrid multi task architecture \cite{Yang2020} on top of BERT and other such methods outlined in \citet{Huang2021}.
As follower counts were found to affect virality, the dataset could be improved further by specifically collecting viral tweets from less popular users.
Additionally, a larger dataset could have been collected to improve the model further (as the dataset was reduced after collection), and data from other social media could be investigated with this model to test adaptability. Lastly, work should be done into incorporating more languages to the dataset to measure how well the models adapt to completely different tweet texts, and a module for media (images, GIFs, videos) could also be introduced through Equation \ref{eq:txt-cls} to determine virality through this new feature type.

\bibliographystyle{ACM-Reference-Format}
\bibliography{sample-base}


\end{document}